\documentclass[10pt,twocolumn,letterpaper]{article}

\usepackage{wacv}              

\usepackage{graphicx}
\usepackage{amsmath}
\usepackage{amssymb}
\usepackage{utfsym}
\usepackage{booktabs}
\usepackage[accsupp]{axessibility}  

\usepackage[pagebackref,breaklinks,colorlinks]{hyperref}

\usepackage[capitalize]{cleveref}
\crefname{section}{Sec.}{Secs.}
\Crefname{section}{Section}{Sections}
\Crefname{table}{Table}{Tables}
\crefname{table}{Tab.}{Tabs.}

\def\wacvPaperID{1036} 
\def\confName{WACV}
\def\confYear{2024}

\begin{document}

\title{GazeGNN: A Gaze-Guided Graph Neural Network for Chest X-ray Classification}

\author{Bin Wang$^{1}$,
Hongyi Pan$^{1}$,
Armstrong Aboah$^{1}$,
Zheyuan Zhang$^{1}$,
Elif Keles$^{1}$,
Drew Torigian$^{2}$,\\
Baris Turkbey$^{3}$,
Elizabeth Krupinski$^{4}$, 
Jayaram Udupa$^{2}$,
Ulas Bagci$^{1}$\\
$^{1}$ Northwestern University,
$^{2}$ University of Pennsylvania,
$^{3}$ National Cancer Institute \\
$^{4}$ Emory University \\
{\tt\small ulas.bagci@northwestern.edu}
}
\maketitle

\begin{abstract}
Eye tracking research is important in computer vision because it can help us understand how humans interact with the visual world. 
Specifically for high-risk applications, such as in medical imaging, eye tracking can help us to comprehend how radiologists and other medical professionals search, analyze, and interpret images for diagnostic and clinical purposes. 
Hence, the application of eye tracking techniques in disease classification has become increasingly popular in recent years. 
Contemporary works usually transform gaze information collected by eye tracking devices into visual attention maps (VAMs) to supervise the learning process. 
However, this is a time-consuming preprocessing step, which stops us from applying eye tracking to radiologists' daily work.
To solve this problem, we propose a novel gaze-guided graph neural network (GNN), GazeGNN, to leverage raw eye-gaze data without being converted into VAMs.
In GazeGNN, to directly integrate eye gaze into image classification, we create a unified representation graph that models both images and gaze pattern information.
With this benefit, we develop a real-time, real-world, end-to-end disease classification algorithm for the first time in the literature.
This achievement demonstrates the practicality and feasibility of integrating real-time eye tracking techniques into the daily work of radiologists.
To our best knowledge, GazeGNN is the first work that adopts GNN to integrate image and eye-gaze data. Our experiments on the public chest X-ray dataset show that our proposed method exhibits the best classification performance compared to existing methods. The code is available at https://github.com/ukaukaaaa/GazeGNN.
\end{abstract}

\section{Introduction}
Image classification has always been a complicated task in the computer vision field.
In recent years, because of the explosive development of machine learning techniques, deep learning-based classification algorithms have been proposed to deal with this challenging task~\cite{lecun1998gradient,krizhevsky2017imagenet,simonyan2014very,he2016deep,ioffe2015batch,wang2022deep}. 
However, compared to the classical natural image datasets such as ImageNet-1k~\cite{deng2009imagenet}, medical image datasets are usually characterized by a relatively limited scale and low signal-to-noise ratio~\cite{chow2016review}, which makes disease classification a more challenging task. 
\begin{figure*}[t]
\centering
{\includegraphics[width=0.83\linewidth]{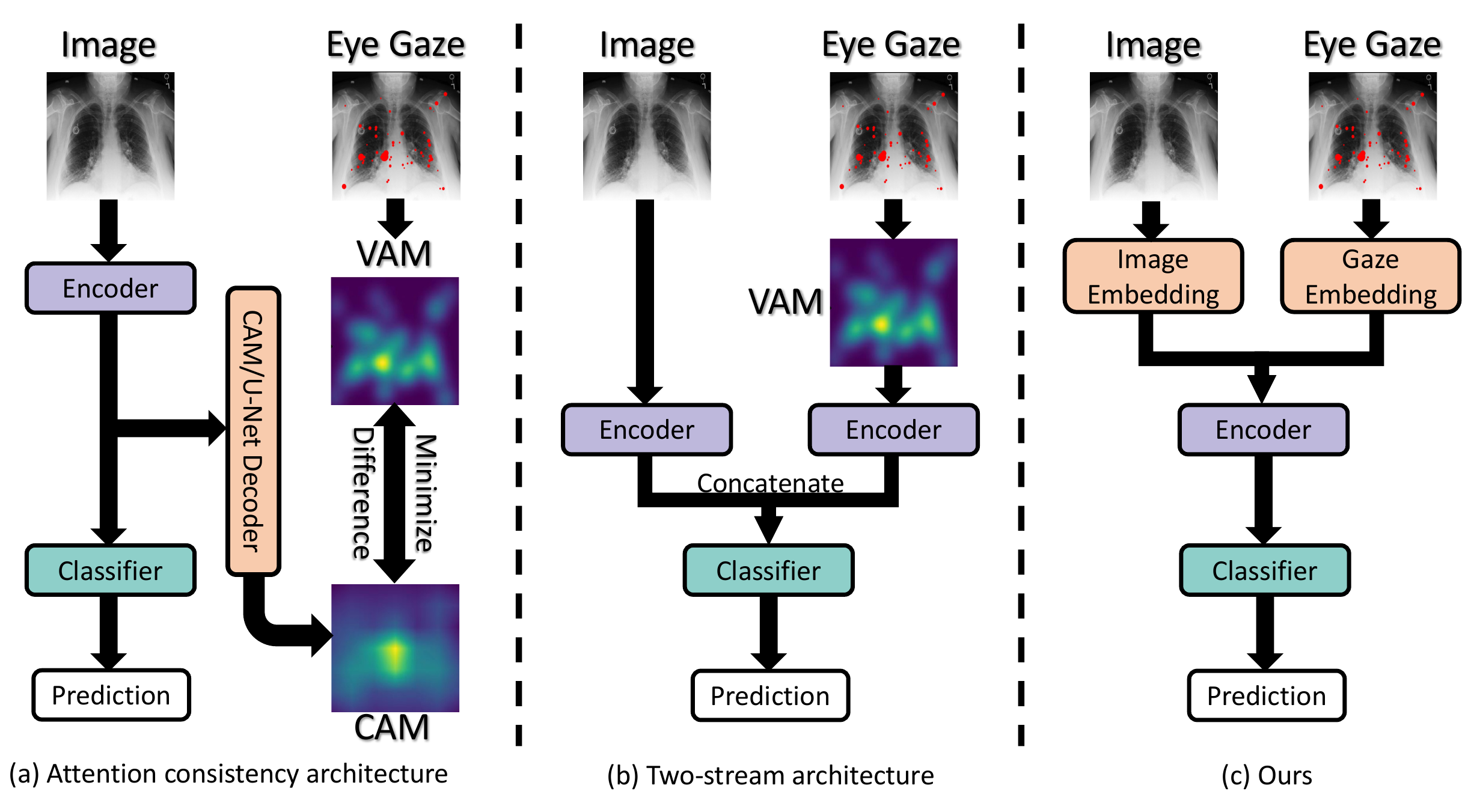}}
\caption{Illustration of our proposed method and other frameworks that integrate eye-gaze information in medical image classification.}
\label{fig: concepts}
\end{figure*}
This problem is particularly evident in chest X-ray classification. It is because chest X-ray has limited soft tissue contrast, containing a variety of complex anatomical structures overlapping in planar (2D) view~\cite{periard1996diagnostic}. Many tissues, such as organs, blood vessels, and muscles, have similar intensity values on the chest X-ray images~\cite{rossmann1970central}. This can easily confuse the deep learning model to distinguish between normal and abnormal tissues, making it difficult to identify the true location of abnormalities accurately.

Therefore, deep learning algorithms encounter difficulties in accurately identifying abnormality based solely on chest X-ray images.
To overcome this challenge, many recent studies have applied eye-tracking techniques to complement the model with prior knowledge of the location of abnormality regions.
Eye-tracking techniques collect eye-gaze data from radiologists during screening procedures~\cite{stember2020integrating,stember2019eye}. This eye-gaze data represents the search pattern of radiologists for tumors or suspicious lesions on the scans. It indicates the location information that radiologists have fixations and saccades on the images during diagnostic screenings. Since these positions are highly likely to hold abnormality and potentially important regions, eye-gaze data can provide extra location information of the disease that is often challenging to be observed from medical images alone. This supplementary information, a high-level attention, can guide the deep learning model to learn the disease feature in an interpretable way. 
Hence, embedding eye-gaze information into diagnostic analysis has become a popular topic in recent years~\cite{bhattacharya2022gazeradar,khosravan2017gaze2segment,khosravan2019collaborative,stember2020integrating}.

The prior mainstream works on this topic can be broadly categorized into two approaches. 
The first one~\cite{bhattacharya2022radiotransformer,wang2022follow,karargyris2021creation,watanabe2022improving, gao2022aligning,zhu2022gaze,saab2021observational,van2023probabilistic} is referred to as the attention consistency architecture, illustrated in Fig.~\ref{fig: concepts}(a). It calculates the attention map based on the model learned by image. At the same time, eye gaze is utilized to supervise the attention map generated by the model. This ensures that the model's attention aligns closely with the attention patterns observed by human experts. 
However, since this architecture only utilizes eye gaze during training as a supervision source and excludes it during testing, there is a potential risk in classification performance and model robustness. This is related to the inherent variability in eye-gaze data. The eye-gaze data can differ significantly from case to case since each radiologist may have their own unique search patterns. 
This individualized nature of eye-gaze data may introduce inconsistencies that complicate the learning process for classification models. 
Therefore, it is challenging to learn a generalized model to capture standardized eye-gaze data patterns for one specific disease. In section \ref{robustsec}, we verify that attention consistency architecture exhibits poor model robustness and has a remarkable performance drop when distribution gaps exist in the data. This motivates us to study other structures to integrate eye-gaze information.

The second approach ~\cite{ma2022rectify,ma2022eye,rong2021human,karargyris2021creation} is known as the two-stream architecture, as depicted in Fig.~\ref{fig: concepts}(b). It consists of two branches dedicated to processing the image and eye gaze information separately. These branches extract features from their respective sources, which are then concatenated and fed into the classification head. In the end, the predicted probabilities of each disease class are achieved.
However, since the eye-gaze data consists of a group of fixation points,
which is not a regular grid or sequence representation, two-stream architecture transforms the eye-gaze data into visual attention maps (VAMs) and then integrates the VAMs with the medical images. It is not ideal for real-world clinical practice because it is time-consuming to generate VAM for each image during inference ($ \sim $10s for each image). There is still a need to prepare all the VAMs in advance before sending them into the network one by one. As a result, this hinders the practical application of eye tracking techniques in the daily clinical workflow.

Therefore, to address the problems of the existing two architectures, we develop a new framework illustrated in Fig.~\ref{fig: concepts}(c).
We consider eye gaze as the model input to enhance the model robustness and directly utilize the raw eye-gaze data without converting it to the VAMs to improve time efficiency.
To bypass the usage of VAM and fully integrate eye gaze with image, we apply a \textit{\textbf{graph}} to model multiple information in a single representation and adopt the Graph Neural Network (GNN) to learn the graph. 
Unlike the nowadays' widely sought Transformer model, GNN is shown to be highly effective even with limited training data, making it a better choice for medical settings~\cite{hamilton2017inductive}.
Additionally, GNN has the advantage of capturing the relational information between different parts of the image according to their semantic and categorical attributes~\cite{han2022vision}. This capability facilitates the learning of relationships between various organs and even the distinction between normal and abnormal regions within the image.

To adapt GNN for disease classification, the image is divided into patches to construct a graph. In the graph, each node stands for a feature fused from three types of information: the location of the patch in the image, the local intensity information of the image patch, and the human attention information from the patch. Respectively, we employ three different embedding techniques to encode the information respectively: (i) positional embedding for the location of the patch, (ii) patch embedding for patch local intensity values extraction, and (iii) gaze embedding for aggregating the fixation time of radiologists on the patch. Then, for each patch, the three embedding features are combined as a single feature vector. Finally, each node is connected to its $k$-nearest neighbors to build the graph. By feeding the graph into a GNN, we obtain the disease classification model.

The major contributions of this work are summarized as: 
\begin{enumerate}

    \item We propose a novel Gaze-guided GNN framework, \textbf{GazeGNN}, which can directly integrate raw eye-gaze data with images, bypassing the need to convert gaze into VAMs. This reduces the inference time of each case from $ \sim $10s to less than 1s, making it the first study that can be applied to real-world clinical practice due to its efficiency and seamless integration.

    \item We leverage the flexibility of a graph network to design a unified graph representation that can encode multiple types of information 
    - the location of the patch in the image, the local intensity information of the image patch, and the human attention information focused on the patch - within a single representation.

    \item Rather than a supervision source, we verify incorporating eye-gaze data as a model input  that can enhance the model's robustness and reduce performance drop in scenarios where distribution gaps exist.
    
    \item By evaluating GazeGNN on a public chest X-ray dataset~\cite{karargyris2021creation}, our proposed method achieves the state-of-the-art performance on the disease classification task. It outperforms the existing strategies that utilize both image and eye-gaze data, from the perspectives of accuracy, robustness, and time efficiency.
\end{enumerate}

\begin{figure*}[t]
\centering
{\includegraphics[width=0.905\linewidth]{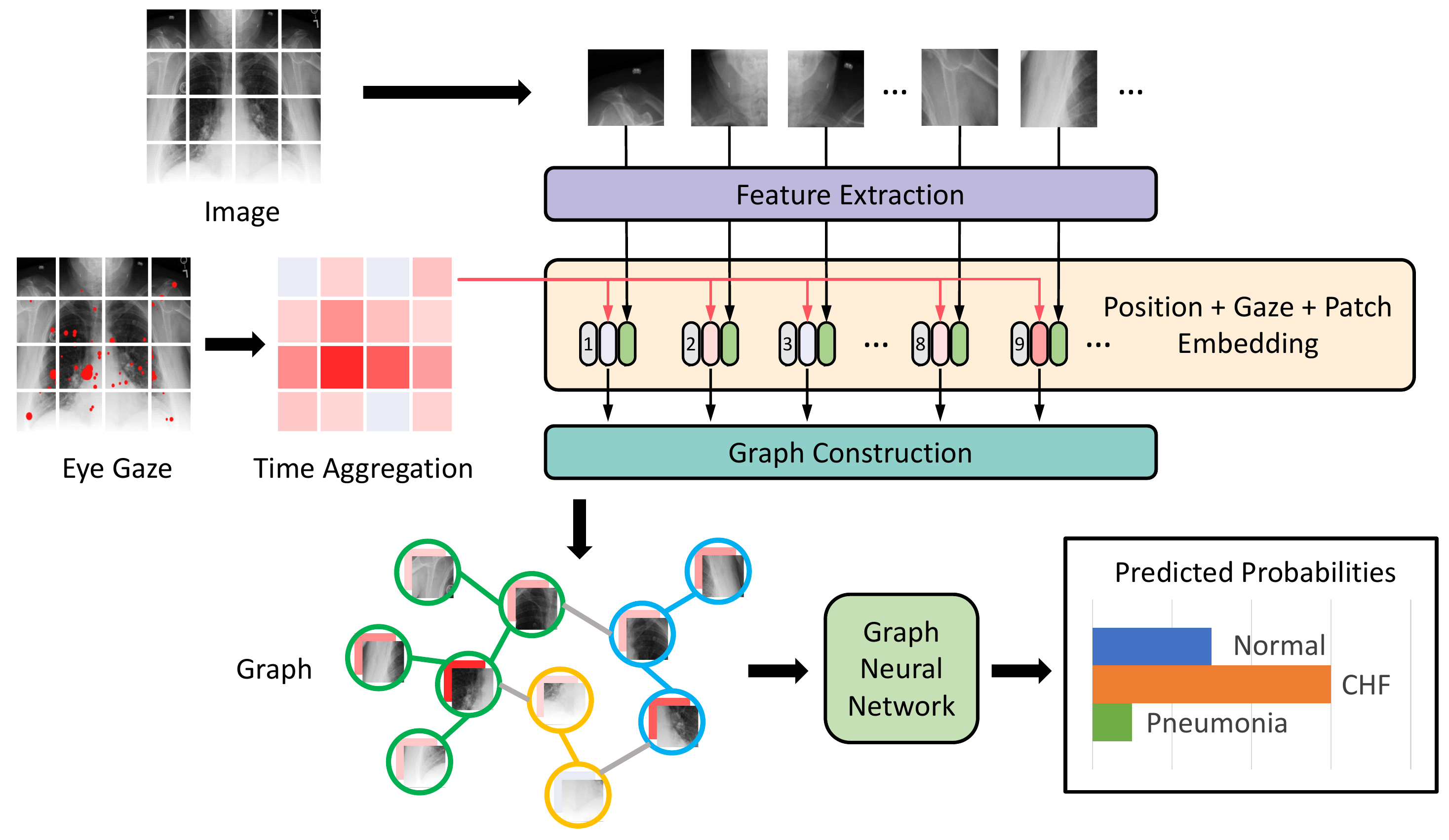}}
\caption{An overview of our proposed GazeGNN framework. It includes a graph construction based on patch, gaze, and position embeddings and a graph neural network for disease classification.}
\label{fig: framework}
\end{figure*}
\section{Related Works}
\subsection{Chest X-ray Classification}
Chest X-ray classification has witnessed significant advances in recent years with the power of large-scale public chest X-ray datasets and advanced machine learning techniques.
Large-scale chest X-ray datasets such as CheXpert~\cite{irvin2019chexpert}, MIMIC-CXR~\cite{johnson2019mimic}, ChestX-ray14~\cite{rajpurkar2017chexnet}, and others~\cite{bustos2020padchest,zhu2013prostate,demner2012design,kermany2018identifying}
have significantly contributed to the model training and evaluation. These datasets provide nearly a million of chest X-ray images (in total) with class annotations, enabling the development of chest X-ray classification algorithms developments. 
On the other side, the development of advanced deep learning algorithms has enhanced the accuracy and performance of chest X-ray classification. Most methods are mainly based on chest X-ray images and propose new network architectures to conduct the analysis~\cite{rajpurkar2017chexnet,islam2017abnormality,baltruschat2019comparison,shin2016learning,guendel2019learning}. Very recently, studies have started to explore the eye-gaze data on chest X-ray classification task~\cite{karargyris2021creation,wang2022follow,van2023probabilistic}. 
Research has shown that the inclusion of additional human expert knowledge via eye-gaze patterns can significantly enhance the accuracy of deep learning models. 

\subsection{Integration of Eye-gaze Data in Medical Image Analysis}
The prior mainstream works commonly transform the eye-gaze data into VAMs.
A VAM is an image that highlights radiologists' attention regions on the corresponding medical image.
Related works can be generally divided into two categories based on their utilization of the VAMs. The first category considers the VAMs as a part of the input for the network models. For example, in~\cite{ma2022rectify,ma2022eye,rong2021human}, authors apply VAMs to process the images and take the processed images as the model input. 
In~\cite{karargyris2021creation}, authors employ a CNN-LSTM hybrid two-stream neural network, where the CNN is used to process the medical images, and the LSTM is used to encode the VAMs. The second category minimizes the difference between VAMs and class activation maps (CAMs)~\cite{bhattacharya2022radiotransformer,wang2022follow} or the difference between VAMs and the attention maps generated by a U-Net decoder~\cite{karargyris2021creation,watanabe2022improving}. 

More recently after the release of Segment Anything Model (SAM) by Meta, a human-computer interaction system, GazeSAM~\cite{wang2023gazesam}, is proposed. Basically, it combines eye tracking technology with SAM and enables users to segment the object they are looking at in real-time, which noticeably proves the possibility of bringing real-time eye gaze integration into routine clinics.


\section{Method}
In this section, we describe the framework of the proposed GazeGNN for the disease classification task. 
As illustrated in Fig.~\ref{fig: framework}, GazeGNN constructs a graph from an image and eye-gaze data.
Each node in the graph is represented as a combination of features through patch, gaze, and position embedding. After graph is constructed, a graph neural network is applied to update and aggregate the information of all the nodes in order to produce a feature representing the whole graph. By performing graph-level classification, we can obtain the predicted class for the input image.

\subsection{Graph Representation}
Our proposed GazeGNN method takes two distinct data types as input: the chest X-ray image and the corresponding eye-gaze information. The image is a regular grid structure data, while eye-gaze information is a group of scatter points that indicates the attention locations of radiologists during their evaluation process. To integrate both types of information effectively, we employ following techniques to embed them into feature vectors to construct a graph accordingly.

\subsubsection{Patch Embedding}
The image input size in this task is $224\times 224$. Therefore, if we treat each pixel as an individual node, there will be 50,176 nodes in the graph. This is an excessive number and makes the GNN training difficult. Instead, we divide the image into multiple $15\times 15$ patches and consider each patch as a node. 

Given an image $\mathcal{I} \in \mathbb{R}^{H \times W}$, we split it into $N$ patches $\mathcal{P} = \{p_1, p_2, ..., p_N\}$, where $p_i \in \mathbb{R}^{S\times S}$ for $i = 1, 2, ..., N$. 
For each patch $p_i$, we extract a feature vector $\mathbf{x}_i^{(I)} \in \mathbb{R}^{D}$ that encodes the local image information, which is:
\begin{equation}\label{Eq: image feature vector}
    \mathbf{x}_i^{(I)} = F(p_i),
\end{equation}
where $F(\cdot)$ is the feature extraction method. In this work, we adopt the overlapping patch embedding method \cite{wang2022pvt} to extract the feature vectors from image patches.

\subsubsection{Gaze Embedding}
Eye-gaze data consists of many scatter points, and each of them means that the radiologists' eyes have concentrated on this location for a moment when they were performing image reading. 
More importantly, eye gaze not only provides the location information but also offers the time duration for each point. 
As illustrated in ``Eye Gaze'' of Fig.~\ref{fig: framework}, there are many red dots with different sizes scattered on the image. A bigger red dot indicates that the radiologist has spent a relatively longer time focusing on the corresponding area.
To maintain consistency with the feature vector defined for a single image patch in Eq.~(\ref{Eq: image feature vector}), we perform time aggregation to get the fixation time for each patch.
\begin{figure*}[htb]
\centering
{\includegraphics[width=0.88\linewidth]{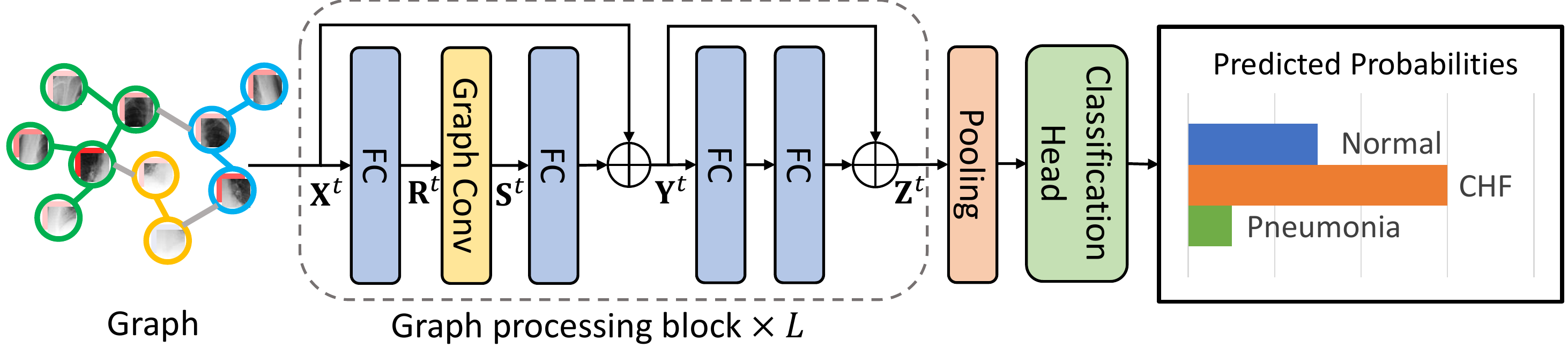}}
\caption{The architecture of the proposed Graph Neural Network (GNN).}
\label{fig: GNN}
\end{figure*}
Assume that there are $Q$ eye-gaze points $g_{(m_1,n_1)}, g_{(m_2,n_2)}, ...,$ $ g_{(m_Q,n_Q)}$, in which $g_{(m_i,n_i)}$ indicates that radiologist's eyes fix at location $(m_i, n_i)$ for $g_{(m_i,n_i)}$ seconds. 
Then, to conduct the time aggregation, we sum up all the eye-gaze points' fixation time in the patch to represent the attention feature of the patch, i.e., for each patch $p_i$, the gaze embedding is:
\begin{equation}
    x_i^{(T)} = \sum_{(m_j, n_j)\in p_i} g_{(m_j, n_j)},
\end{equation}
where $i\in [1,N]$ and $j\in [1,Q]$. Next, we replicate the scalar $x_i^{(T)}$ to the vector $\mathbf{x}_i^{(T)}\in \mathbb{R}^{D}$ for feature fusion. 

\subsubsection{Position Embedding}
During the graph processing in GNN, the features are treated as unordered nodes. To keep the positional information in the original image, we adopt the position embedding method from~\cite{han2022vision}, which contains two steps. The first step is to add a learnable absolute positional encoding vector $\mathbf{e}_i \in \mathbb{R}^D$ to the feature vector $\left(\mathbf{x}_i^{(I)} + \mathbf{x}_i^{(T)}\right)$. In the second step, we calculate the relative positional distance between nodes as $\mathbf{e}_i^T\mathbf{e}_j$, and this distance is used to determine the neighbors of a given node in the $k$-nearest neighbors algorithm for the graph construction.

\subsubsection{Graph Construction}
With patch, gaze, and position embeddings, the graph node feature vector $\mathbf{x}_i$ is elaborated as:
\begin{equation}\label{eq: feature fusion}
    \mathbf{x}_i = \mathbf{x}_i^{(I)} + \mathbf{x}_i^{(T)} + \mathbf{e}_i,
\end{equation}
and these features represent the vertices $\mathcal{V} = \{\mathbf{x}_1, \mathbf{x}_2, ... \mathbf{x}_N\}$.
By calculating the $k$-nearest neighbors, the edges of the graph are defined as
\begin{equation}
    \mathcal{E} = \{(\mathbf{x}_i, \mathbf{x}_j) \mid \mathbf{x}_j \in K(\mathbf{x}_i)\},
\end{equation}
where $K(\mathbf{x}_i)$ represents the $k$-nearest neighbors of $\mathbf{x}_i$. 
In this way, a graph $G=\{ \mathcal{V}, \mathcal{E} \}$ is constructed.

\subsection{Graph Neural Network (GNN)}
As illustrated in Fig.~\ref{fig: GNN}, the graph neural network consists of $L$ graph processing blocks~\cite{han2022vision}, an average pooling layer, and a graph classification head. 
Graph processing block consists of multiple fully-connected (FC) layers and a graph convolutional layer~\cite{li2019deepgcns}.

Suppose the graph is represented as $N$ $D$-dimension feature vectors.
Given an input graph $\mathbf{X}^t= \left[\mathbf{x}_1^t, \mathbf{x}_2^t, ..., \mathbf{x}_N^t\right]\in \mathbb{R}^{N\times D}$ at block $t$, a graph processing block outputs $\mathbf{Z}^t\in \mathbb{R}^{N\times D}$ as
\begin{equation}
    \mathbf{Y}^t = \Psi_2\left (\Phi \left( \Psi_1 \left( \mathbf{X}^t\right) \right)\right) + \mathbf{X}^t,
\end{equation}
\begin{equation}
    \mathbf{Z}^t = \Psi_4\left( \Psi_3\left ( \mathbf{Y}^t\right)\right) + \mathbf{Y}^t,
\end{equation}
where $\Phi$ denotes the graph convolution operation and $\Psi$ indicates FC layer. Here, we ignore the activation and batch normalization layers. Let $\mathbf{Y}^t\in \mathbb{R}^{N\times D}$ stand for the intermediate output after the first shortcut connection and $\mathbf{R}^{t}=\Psi_1(\mathbf{X}^{t})$ stand for the input of graph convolutional layer. 
The graph convolution $\mathbf{S}^{t}=\Phi(\mathbf{R}^{t})$ is defined as
\begin{equation}
    \mathbf{s}_i^{t}\ = \mathbf{W}\cdot\max\left( \left\{\mathbf{r}_i^{t}-\mathbf{r}_j^{t} \mid j \in K\left(\mathbf{r}_i^{t}\right)\right\}\right),
\end{equation}
where $\mathbf{S}^t= \left[\mathbf{s}_1^t, \mathbf{s}_2^t, ..., \mathbf{s}_N^t\right]\in \mathbb{R}^{N\times D}$ and $\mathbf{R}^t= \left[\mathbf{r}_1^t, \mathbf{r}_2^t, ..., \mathbf{r}_N^t\right]\in \mathbb{R}^{N\times D}$. $\mathbf{W}$ is a trainable weight matrix to update the feature for the node. The max term is the aggregation function that aggregates features from the $i$-th node's neighbors.
Therefore, graph convolution aggregates node neighbors' feature information and updates it into the node feature. In the final step, the classification head is designed as a fully-connected layer with the softmax function. It outputs the predicted probability of each category.

\section{Experiments}
Our experiments are implemented on a workstation with an Intel Xeon W-2255 CPU and an NVIDIA RTX 3090 GPU using PyTorch.  
We train GazeGNN using AdamW optimizer~\cite{loshchilov2018decoupled} with the learning rate of 0.0001 and the batch size of 32. The checkpoint model with the best testing accuracy is saved during the training. Cross-entropy loss is used as the classification loss function. 
In the following experiments, we adopt \cite{karargyris2021creation} as the implementation of two-stream architecture and \cite{wang2022follow} as the implementation of attention consistency architecture.
\begin{table*}[t]
\centering
\caption{Classification results on the Chest X-Ray dataset~\cite{karargyris2021creation}.}\label{tab: Classification results}
\begin{tabular}{l|c|c|c|c|c|c|c|c}
\hline
Method& Accuracy&\multicolumn{4}{|c|}{AUC}& Precision & Recall & F1-Score\\
& &Normal&CHF&Pneumonia&Average &  &  & \\
\hline
Temporal Model~\cite{karargyris2021creation} &-&0.890&0.850&0.680&0.810&-&-&-\\
U-Net+Gaze~\cite{karargyris2021creation}&-&0.910&0.890&0.790&0.870&-&-&-\\
DenseNet121+Gaze~\cite{van2023probabilistic}&-&-&-&-&0.836&-&-&0.270\\
GazeMTL~\cite{saab2021observational}&78.50\%&0.915&0.913&0.833&0.887&0.786&0.781&0.779\\
IAA~\cite{gao2022aligning}&78.50\%&0.922&0.902&0.875&0.900&0.780&0.774&0.776\\
EffNet+GG-CAM~\cite{zhu2022gaze}&77.57\%&0.906&0.914&0.843&0.888&0.770&0.772&0.770\\
\hline
\textbf{GazeGNN}& \textbf{83.18\%} & \textbf{0.938}&\textbf{0.916}  &\textbf{0.914} &\textbf{0.923}&\textbf{0.839}&\textbf{0.821}&\textbf{0.823}\\
\hline
\end{tabular}
\end{table*}
\subsection{Dataset Preparation}
The experiments in this paper are carried out on a public chest X-ray dataset~\cite{karargyris2021creation}, which contains 1083 cases from the MIMIC-CXR dataset~\cite{johnson2019mimic}. 
For each case, a gray-scaled X-ray image with the size of around $3000\times3000$, eye-gaze data, and ground-truth classification labels are provided. 
These cases are classified into 3 categories: Normal, Congestive Heart Failure (CHF), and Pneumonia. 
For the comparison experiments, we generate the static VAMs from the eye-gaze data using the data post-processing method as described in~\cite{karargyris2021creation}.
The model performance is evaluated through multiple metrics, including accuracy, the area under the receiver operating characteristic curve (AUC), precision, recall, and F1-score. The higher these metrics are, the better the model is. 
For all the experiments, we apply the same data augmentation techniques, including random resize crop into $224\times224$, random horizontal flip, and random rotation by up to $5^{\circ}$.

\subsection{Improving Disease Classification Accuracy}

We compare GazeGNN with the state-of-the-art methods, including temporal model~\cite{karargyris2021creation}, U-Net+Gaze model~\cite{karargyris2021creation}, and DenseNet121-based model~\cite{van2023probabilistic}. These methods adopt the official training and test datasets, so we directly include their reported results in this paper.
We also compare GazeGNN with some other gaze-guided methods, which have not been validated on this dataset yet, or have used this dataset but did not follow the official splitting strategy. These methods include GazeMTL~\cite{saab2021observational}, IAA~\cite{gao2022aligning}, and EffNet+GG-CAM~\cite{zhu2022gaze}. To make the comparison fair, we train these methods under the same setting as GazeGNN.

\begin{figure}[t]
\centering
\subfloat[GazeGNN.]
{\includegraphics[width=0.5\linewidth]{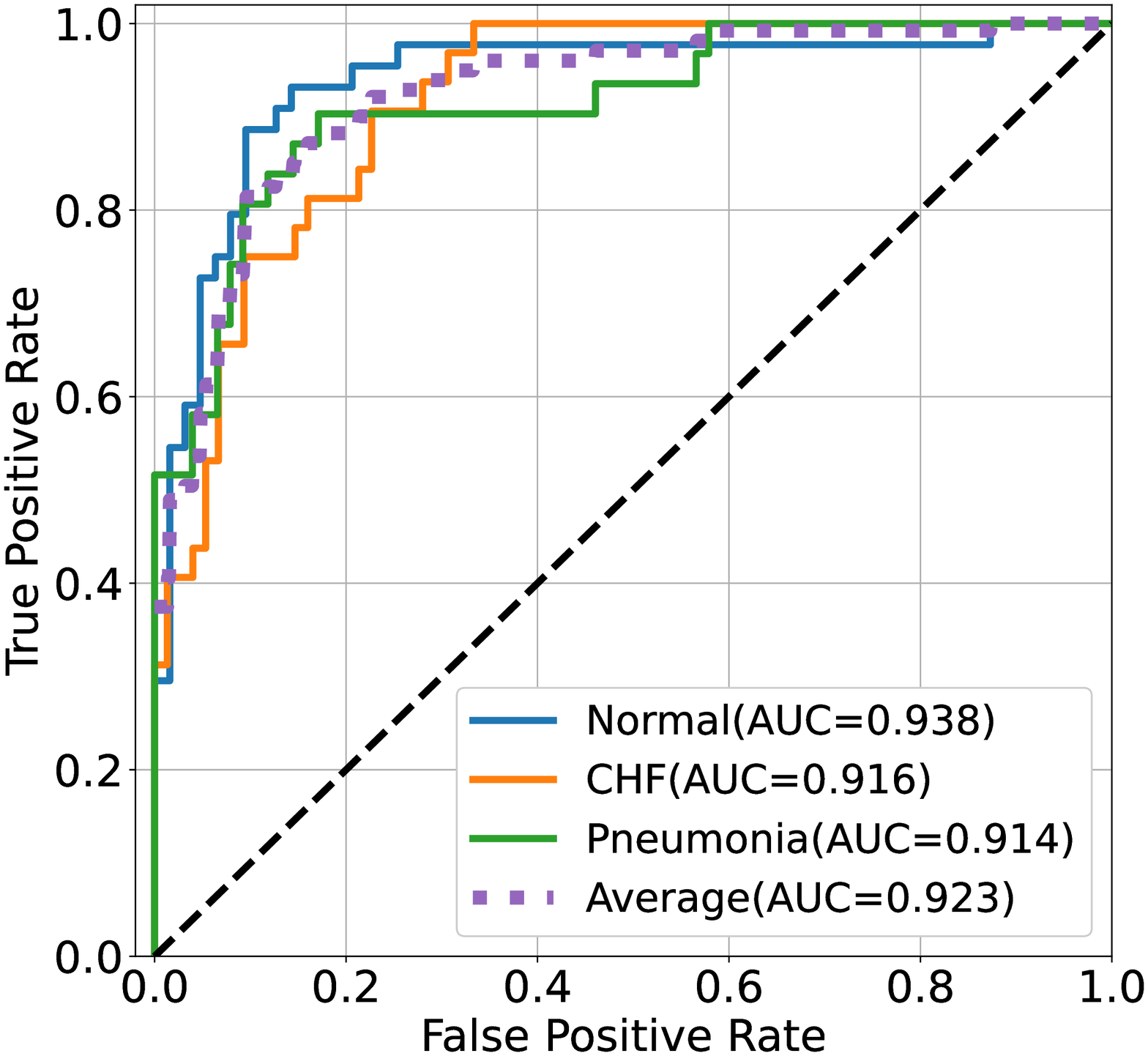}}
\subfloat[IAA~\cite{gao2022aligning}.]{\includegraphics[width=0.5\linewidth]{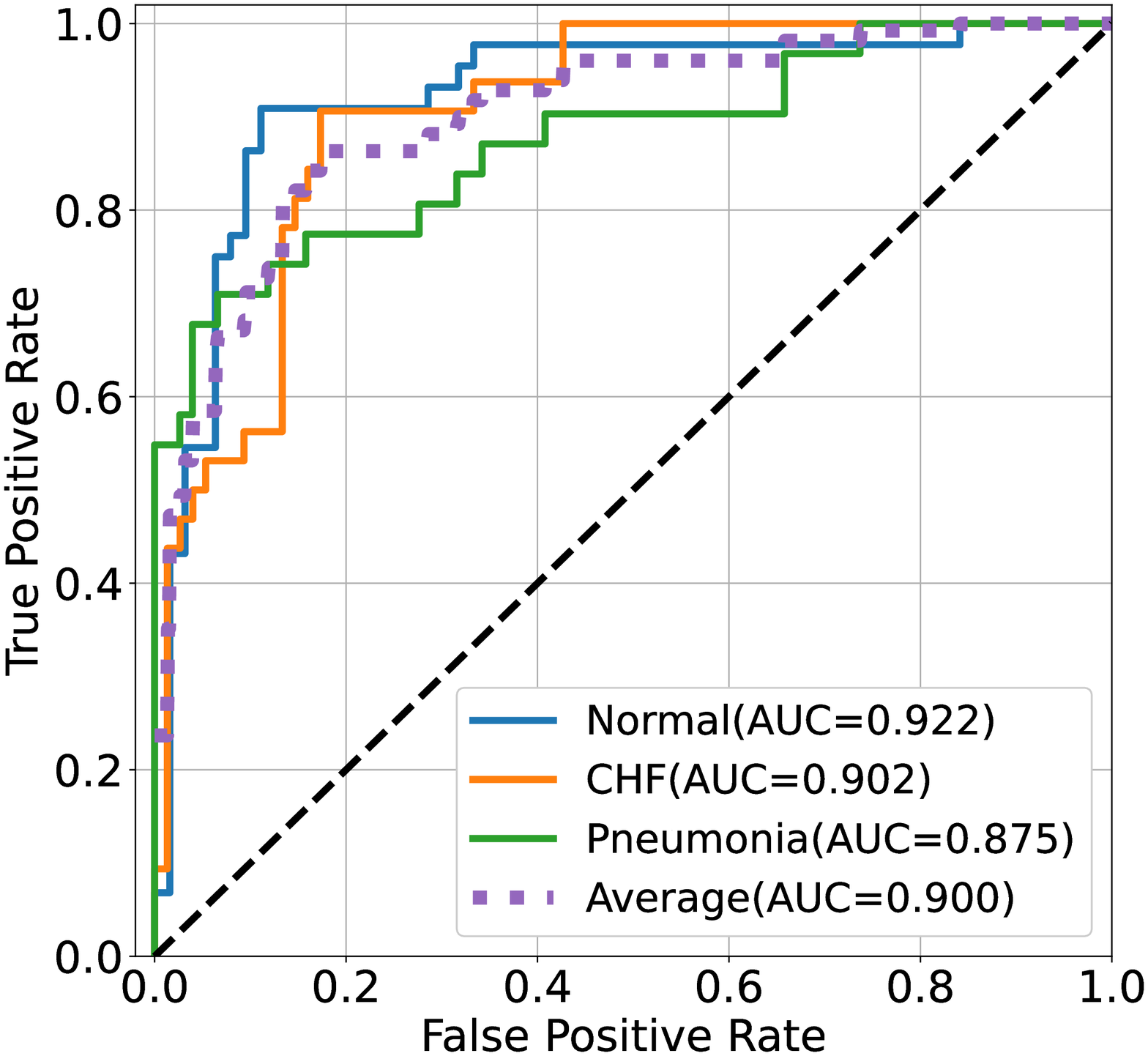}}
\caption{Comparisons of ROC curves and AUC scores}\label{fig: ROC}
\end{figure}

The quantitative results are summarized in Table~\ref{tab: Classification results}. Although we primarily compare the accuracy metric in this work because we save the checkpoint models with the best testing accuracy, it is noted that the proposed GazeGNN still achieves the best performance on all the evaluation metrics. 
Moreover, Fig.~\ref{fig: ROC} shows receiver operating characteristic (ROC) curves of the comparison method with the best average AUC and our GazeGNN. The ROC curves of other compared methods are presented in the supplemental materials.

\subsection{Improving Inference Speed}
Eye-gaze data is composed of a group of scatter points, indicating the location coordinates of the radiologists' gaze on the medical image. It is not a regular grid or sequential data format. To align the eye-gaze data with the medical image, existing methods typically transform the eye-gaze into the VAMs for training purposes. The generation of VAM for each image can be a time-consuming process. There are two approaches to accomplish this step. One method is to apply a Gaussian distribution to each eye-gaze point and aggregate the individual distributions to obtain the final VAM. The other approach is to apply a Gaussian filter kernel to smooth the eye-gaze intensity value (duration time on a certain image location) on the whole image. Due to the large size of chest X-rays (approximately 2500x3000) and the considerable number of eye-gaze points, generating VAMs for each image requires substantial time. Consequently, existing methods often pre-generate all VAMs in advance before training or inference. This is not ideal when we want to integrate the eye-gaze into the radiologists' daily work.
\begin{table}[h]
\caption{Comparison of inference speed.}\label{tab: speed}
\centering
\resizebox{1\columnwidth}{!}{
\begin{tabular}{l|c|c}
\hline
Method&Gaze&Inference Time\\
\hline
GazeGNN& \checkmark&\textbf{0.353s}\\
Two-stream Architecture& \checkmark&9.246s \\
Attention Consistency Architecture& \usym{2717}&0.294s \\
\hline
\end{tabular}
}
\end{table}
\begin{table*}[t]
\centering
\caption{Performance comparison of our method across different backbones.}\label{tab: backbone}
\begin{tabular}{l|c|c|c|c|c|c|c|c}
\hline
Backbone& Accuracy&\multicolumn{4}{|c|}{AUC}& Precision & Recall & F1-Score\\
& &Normal&CHF&Pneumonia&Average &  &  & \\
\hline

DenseNet121~\cite{huang2017densely}&71.03\%&0.903&0.855&0.620&0.793&0.696&0.689&0.689\\
ResNet18~\cite{he2016deep}&71.96\%&0.906&0.820&0.687&0.804&0.706&0.706&0.705\\
ResNet50~\cite{he2016deep}&70.09\%&0.898&0.818&0.663&0.793&0.685&0.685&0.684\\
ResNet101~\cite{he2016deep}&71.03\%&0.852&0.862&0.756&0.823&0.703&0.705&0.703\\
Swin-T~\cite{liu2021swin}&77.57\%&0.925&0.898&0.732&0.852&0.762&0.760&0.755\\
Swin-S~\cite{liu2021swin}&74.77\%&0.911&0.873&0.728&0.837&0.733&0.735&0.733\\
Swin-B~\cite{liu2021swin}&76.64\%&0.907&0.880&0.770&0.852&0.771&0.754&0.748\\
\hline
\textbf{GNN}& \textbf{83.18\%} & \textbf{0.938}&\textbf{0.916}  &\textbf{0.914} &\textbf{0.923}&\textbf{0.839}&\textbf{0.821}&\textbf{0.823}\\
\hline
\end{tabular}
\end{table*}
In our method, on the other hand, we bypass the process of generating VAM and propose a novel technique, called time aggregation with gaze embedding, to conduct eye-gaze integration. Due to the simple calculation inside the time aggregation, we significantly reduce the inference time, as shown in Table~\ref{tab: speed}. We compare the inference speed of our method and the current two mainstream architectures.  We test on 100 cases and calculate the average processing time as the inference time. For attention consistency architecture, a Gaussian filter kernel, with standard deviation $\sigma=150$, is applied to generate the VAM for each case.

From the result shown in Table.~\ref{tab: speed}, we can find that two-stream architecture takes the longest inference time, around 10 seconds. This is mainly due to the time-consuming process of VAM generation. It is worth noting the GazeGNN obtains comparable inference time as attention consistency architecture. The attention consistency architecture does not require gaze input in the inference stage, while GazeGNN involves the eye-gaze. This demonstrates the efficiency of eye-gaze integration in our architecture, which points out the feasibility to bring real-time eye-tracking techniques into the radiology rooms.

\subsection{Improving Model Robustness}\label{robustsec}
\begin{table}[h]
\caption{Comparison of performance drop when testing on the dataset with distribution shift.}\label{tab: robust}
\centering
\resizebox{\columnwidth}{!}{
\begin{tabular}{l|c|c|c|c|c}
\hline
Method&\multicolumn{5}{|c}{Performance Drop $\downarrow$}\\
&Accuracy&Precision&Recall&F1-Score & Average AUC\\
\hline
GazeGNN& \textbf{2.78\%}&\textbf{1.10\%}&\textbf{2.87\%}&\textbf{3.97\%}&\textbf{0.20\%}\\
ACA&13.79\%&15.30\%&15.63\%&18.38\%&4.86\% \\
\hline
\end{tabular}
}
\end{table}
In attention consistency architecture, the eye-gaze data is considered a supervision source during training, as illustrated in Fig.~\ref{fig: concepts}. The inference stage of attention consistency architecture does not involve eye-gaze information. This requires the model to learn the eye-gaze pattern for certain diseases. However, the eye-gaze data is different case by case and each radiologist has his own search patterns when doing image reading. Further, even for the same radiologist's second time reading of the same scan may show differences in eye-gaze patterns. Therefore, learning standardized eye-gaze data patterns for a specific disease is challenging, and likely not a generalizable model.

To fully utilize the power of eye-gaze information, we postulate that the model should incorporate gaze input in the inference stage. In this way, when encountering new data that exhibits a distribution shift from the original training dataset,  we can still leverage the eye-gaze data to provide the model with the location information of the potential abnormality. To prove this assumption, we introduce random noise to the testing dataset, creating a distribution gap from the original training dataset. We then evaluate our method and attention consistency architecture (ACA) on the original and noisy testing datasets. Based on the results presented in Table \ref{tab: robust}, it is evident that the attention consistency architecture exhibits a larger performance drop compared to our proposed method, validating our previous assumption. 

\subsection{Effectiveness of GNN}
After combining the position, gaze, and patch embedding, we obtain a single feature that represents both image and eye gaze. In this work, the feature is used to construct a graph and processed by GNN. But it also works for other backbone architectures. We employ strong backbone networks, including DenseNet, ResNet, and Swin Transformer, and compare the performance with GNN.

The performance of our method across different backbones is shown in Table \ref{tab: backbone}. The Transformer backbone does not exhibit the best performance. This might be because it suffers from limited data. In addition, we see that our method with GNN achieves the best results over all the evaluation metrics. This can be attributed to two key factors. 
Firstly, unlike the Transformer model, GNN demonstrates remarkable effectiveness even when presented with limited training data.  
The other reason is that GNN can capture and comprehend the intricate relationships between patches through graph learning.

\subsection{Ablation Study of Gaze Usage}

\begin{figure*}[t]
\centering
{\includegraphics[width=0.74\linewidth]{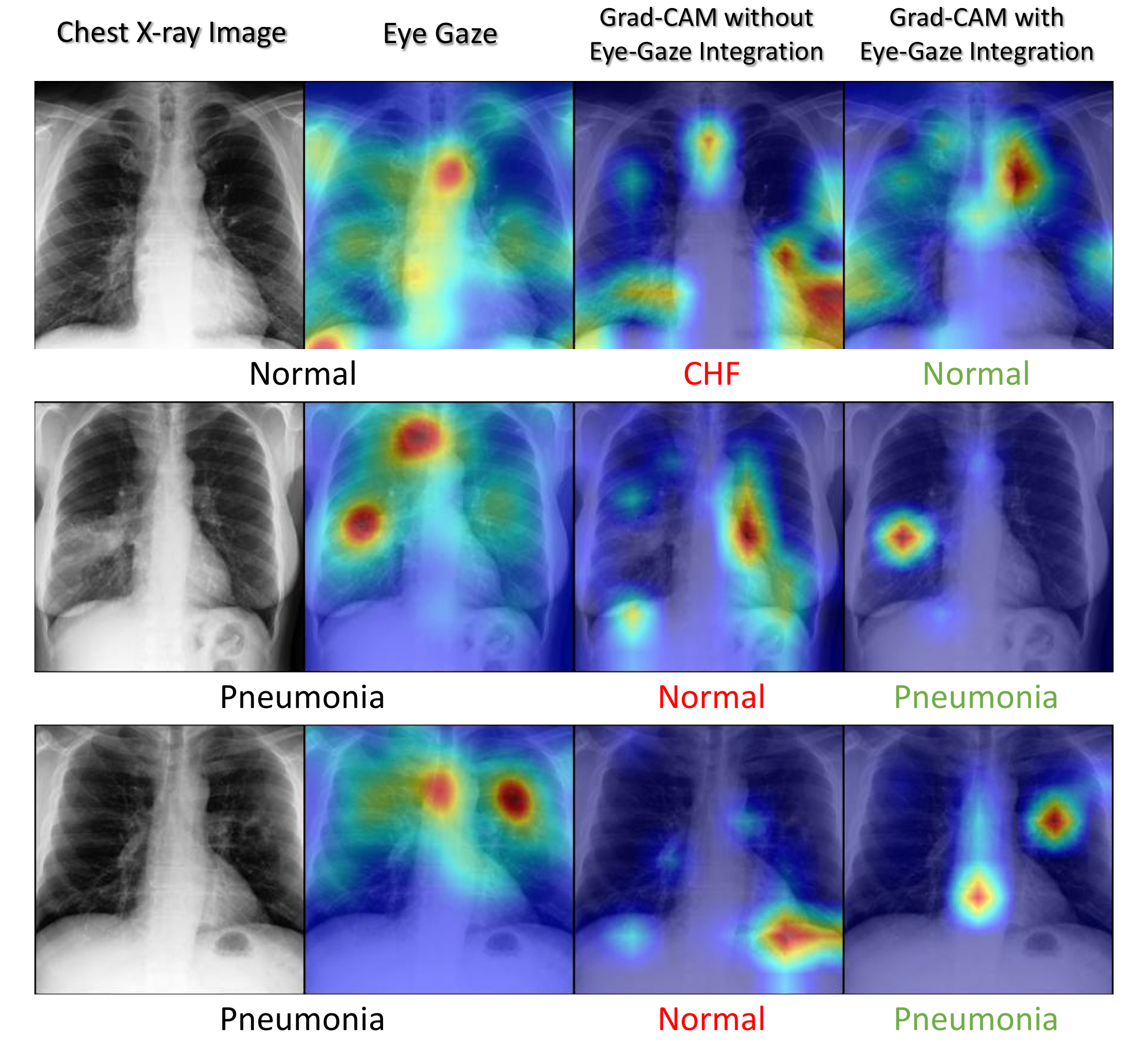}}
\caption{Gaze map and Grad-CAM based attention maps with and without eye-gaze data are shown. Under the images, original label of the chest X-ray is represented by the black color, while the red and green labels indicate incorrect and correct model predictions, respectively.}
\label{fig: visualize}
\end{figure*}
To study the effectiveness of the gaze information, we remove the gaze embedding and only fuse the features from patch embedding and position embedding. In this way, gaze information is not used.
The comparison is presented in Table~\ref{tab: ablation study} and supplementary. 
Without the gaze information, accuracy, average AUC, precision, recall, and F1-score all descend. This validates the assumption that introducing the eye-gaze data can improve classification performance.
It is noted that even without gaze embedding, our obtained accuracy is higher than 80\% and the average AUC is higher than 0.900, superior to most gaze-guided state-of-the-art methods. This is because the proposed graph representation is powerful enough to help the model recognize the image.
\begin{table}[htbp]
\caption{Ablation study on GazeGNN with/without the eye-gaze information.}\label{tab: ablation study}
\centering
\resizebox{\columnwidth}{!}{
\begin{tabular}{l|c|c|c|c|c}
\hline
Gaze&Accuracy&Average AUC&Precision& Recall&F1-Score\\
\hline
\checkmark& \textbf{83.18\%}&\textbf{0.923}&\textbf{0.839}&\textbf{0.821}&\textbf{0.823}\\
\usym{2717}& 80.37\% &0.910&0.800&0.805&0.801\\
\hline
\end{tabular}
}
\end{table}
In addition, we visualize the model's intermediate features to show the power of eye-gaze integration. We use Grad-CAM~\cite{selvaraju2017grad} to generate the attention map from the trained model. From Fig.~\ref{fig: visualize}, it is observed that before the eye-gaze integration, the model fails to focus on the abnormal regions, resulting in incorrect classification decisions. However, when eye-gaze is introduced, the model's attention shifts to the regions highlighted by radiologists. This indicates the guidance of eye-gaze enhances the model's capability to achieve more accurate abnormality localization.


\section{Conclusion}

In this study, we propose a novel gaze-guided graph neural network, GazeGNN, to perform the disease classification task. With the flexibility of graph representation, GazeGNN can utilize the raw eye-gaze information directly by embedding it with the image patch and the position information into the graph nodes. Therefore, this method avoids generating the VAMs that are required in mainstream gaze-guided methods. With this benefit, we develop a real-time, end-to-end disease classification algorithm without preparing the visual attention maps in advance. We show that GazeGNN can produce a significantly better performance than existing methods under the same training strategy. This proves the feasibility of bringing real-time eye tracking techniques to radiologists' daily work. 
\section{Acknowledgment}
This study is supported by NIH R01-CA246704, R01-CA240639, R15-EB030356, R03-EB032943, U01-DK127384-02S1, and U01-CA268808. Sincerely thank Mingfu Liang from Northwestern University for the constructive suggestions on this paper.

{ 
\small
\bibliographystyle{ieee_fullname}
\bibliography{main}
}

\end{document}